\documentclass[conference]{IEEEtran}
\IEEEoverridecommandlockouts
\usepackage{cite}
\usepackage{amsmath,amssymb,amsfonts}
\usepackage{algorithmic}
\usepackage{graphicx}
\usepackage{textcomp}
\usepackage{xcolor}
\usepackage{multirow,multicol}
\usepackage{diagbox}
\usepackage{array}
\usepackage{setspace}
\usepackage{subfigure}
\usepackage[breaklinks=true,bookmarks=false]{hyperref}
\def\BibTeX{{\rm B\kern-.05em{\sc i\kern-.025em b}\kern-.08em
    T\kern-.1667em\lower.7ex\hbox{E}\kern-.125emX}}

\newcommand{\RomanNumeralCaps}[1]
    {\MakeUppercase{\romannumeral #1}}

\makeatletter
\newcommand{\linebreakand}{%
  \end{@IEEEauthorhalign}
  \hfill\mbox{}\par
  \mbox{}\hfill\begin{@IEEEauthorhalign}
}
\makeatother

\begin{document}

\title{ICE-GAN: Identity-aware and Capsule-Enhanced GAN with Graph-based Reasoning for Micro-Expression Recognition and Synthesis\\
}


\author{\IEEEauthorblockN{Jianhui Yu\textsuperscript{1},
Chaoyi Zhang\textsuperscript{1}, Yang Song\textsuperscript{2},
Weidong Cai\textsuperscript{1}}
\IEEEauthorblockA{\textsuperscript{1}School of Computer Science, University of Sydney, Australia
\\\{jianhui.yu, chaoyi.zhang, tom.cai\}@sydney.edu.au}
\IEEEauthorblockA{\textsuperscript{2}School of Computer Science and Engineering, University of New South Wales, Australia
\\yang.song1@unsw.edu.au}}
\maketitle

\begin{abstract}
Micro-expressions are reflections of people’s true feelings and motives, which attract an increasing number of researchers into the study of automatic facial micro-expression recognition. The short detection window, the subtle facial muscle movements, and the limited training samples make micro-expression recognition challenging. To this end, we propose a novel Identity-aware and Capsule-Enhanced Generative Adversarial Network with graph-based reasoning (ICE-GAN), introducing micro-expression synthesis as an auxiliary task to assist recognition. The generator produces synthetic faces with controllable micro-expressions and identity-aware features, whose long-ranged dependencies are captured through the graph reasoning module (GRM), and the discriminator detects the image authenticity and expression classes. Our ICE-GAN was evaluated on Micro-Expression Grand Challenge 2019 (MEGC2019) with a significant improvement (12.9\%) over the winner and surpassed other state-of-the-art methods.
\end{abstract}


\section{Introduction}
Micro-expressions (MEs) display unconscious feelings that can be hardly perceived by untrained observers, making it a challenging pattern recognition task.
However, it has attracted an increasing number of researchers into the study of micro-expression recognition (MER) due to its practical applications in areas such as \cite{intro1} and \cite{intro2}.
There have already been many successful works proposed for general expression recognition, but the domain of MER is poorly-explored mainly because no large-scale ME datasets are available to support extensive MER studies.
Therefore, we introduce micro-expression synthesis (MES) as an auxiliary task to intentionally increase the number of ME samples.

Before the application of deep learning algorithms, hand-engineered methods for recognizing MEs are used according to \cite{intro1}.
Facial Action Coding System (FACS) \cite{6} is applied to recognize facial expressions, which pays attention to muscles that produce the expressions and measures the movement with the help of action units (AUs).
Two systems are further developed: the Micro Expression Training Tool (METT) and Subtle Expression Training Tool (SETT) \cite{7}.
However, the best classification accuracy achieved by METT/SETT is still low because the result is heavily affected by humans, making the detection unconvincing and unstable.
Local binary pattern (LBP) and local quantized pattern (LQP) are later developed \cite{intro1}, and LBP with three orthogonal planes (LBP-TOP \cite{8}) has shown superiority in processing facial images.
However, these geometry-based methods rely heavily on the proposed images and can be easily affected by global changes.

With the development of deep learning technologies, especially convolutional neural networks (CNNs), lots of works have been proposed based on data-driven approaches for MER \cite{25,18,32}.
ELRCN \cite{25} adopts the model architecture of \cite{26} with enriched features to capture subtle facial movements from frame sequences.
Moreover, methods utilize the optical flow to enhance the model performance.
STSTNet \cite{17} extracts three optical features for lightweight network construction.
Dual-Inception \cite{dualin} learns the facial MEs with the help of horizontal and vertical optical flow components.
ME-Recognizer \cite{18} obtains optical features encode the subtle face motion with domain adaptation, which achieves the $1^{st}$ place in MEGC2019 \cite{see2019megc}.
However, CNN-based methods are invariant to translations and do not encode positional relations, and orientation information of different facial entities is also ignored. Furthermore, due to the quantity limit of ME samples, the model performance is heavily constrained.

Capsule network \cite{19} is translation equivariant, which is adopted for MER in \cite{32} to distinguish meaningful facial patterns.
In addition, the size limitation of current ME datasets (i.e., \textit{SMIC} \cite{smic}, \textit{CASME \RomanNumeralCaps{2}} \cite{casme2}, and \textit{SAMM} \cite{samm}) puts a constraint on deep learning methods that can hardly derive benefits from small-scale datasets.
However, this can be resolved by adopting generative adversarial network (GAN) \cite{20} and the variants \cite{cgan,38,33}, which have shown significant generative capabilities in various vision application fields \cite{40,wang2018vid2vid}, so that GANs can naturally help to expand the size of micro-expression datasets.

Relation learning can be crucial for the general visual analysis \cite{glore}, and capturing global relations during the training process is quite beneficial \cite{wang2020micro}.
We propose to learn relations globally on feature channels to capture the long-range channel interdependencies as in \cite{senet}, with the help of self-attention mechanism \cite{attention}.
Moreover, graph convolution networks (GCNs) \cite{gcn} are proposed and adopted initially by \cite{zhong2019graph} for feature representation and textual information learning, where additional information such as AUs is needed to construct the graph. However, we propose to construct graphs based on channel features, so the amount of needed information is heavily reduced and global relations between different facial attributes can be learned, where node features and edge features are learned during training.

The main contributions of our work can be summarized as 3-fold:
(1) We design an identity-aware generator that produces controllable expressions based on side information and generates realistic faces with a graph reasoning module, which models channel interdependencies via attention mechanism to fully explore global relations between facial encodings on local responses.
(2) We introduce a multi-tasking capsule-enhanced discriminator to distinguish image authenticity and predict micro-expression labels, with position-insensitive issues alleviated by the capsule-based algorithm to improve the MER accuracy.
(3) We thereby develop an ICE-GAN framework for identity-aware MES with graph-based reasoning and capsule-enhanced MER, which outperforms the winner of the MEGC2019 Challenge benchmark and recently proposed methods.

\section{Related Work}
\subsection{Generative Adversarial Network}
Existing databases for MER constrain the development of data-driven deep learning methods because overfitting is considered a major problem.
To enlarge the current dataset, GANs can generate unseen images by modeling the training data distribution, where distribution mappings are learned by the generator and the authenticity of inputs is determined by the discriminator.
So far there have been tremendous extensions of GANs \cite{33,40,cgan}, and applications of GANs to MEs are also explored \cite{latestMER,xie2020assisted}.
For instance, optical flow images are generated via GAN to enlarge the database in \cite{latestMER}, and real facial images are produced based on AU intensity in \cite{xie2020assisted}.
However, our method is inspired by Auxiliary Classiﬁer GAN (ACGAN) \cite{38} to generate unseen MEs based on image features with controllable categories, which can be of high quality and high discriminability.


\subsection{Graph-based Reasoning}
Global relations between different nodes are initially modeled by conditional random fields \cite{krahenbuhl2011efficient}.
Recently, graph convolutional networks (GCNs) \cite{gcn} are proposed and adopted in \cite{glore,GINet,wang2021single}, and \cite{liang2018symbolic}, where relational information is reasoned over graphs, and contextual interactions over textual or visual contents are captured.
Besides, graph-based reasoning methods are applied in the study field of MEs. \cite{xie2020assisted,MER-GCN,lei2020novel} facilitate the learning of subtle facial movements from graphs that are established on AU nodes, and structural relations are captured for geometric reasoning.
However, instead of directly constructing graphs over raw data inputs, we propose to explore global relations over intermediate feature channels for representative relation learning.
Inspired by SENet \cite{senet} which applies attention mechanism \cite{attention} to model the interactions between input features and channel features where spatial information is pooled, we employ self-attention within feature channels to save low-level spatial features, resulting in an attentive mapping that contains dependencies between facial attributes.
Moreover, different from \cite{xie2020assisted} which implements MER with supplementary information such as AUs, our work can directly learn the relational information of different face regions, reducing the amount of information required.




\subsection{Capsule Network}
Different from conventional CNNs that are translation invariant, the capsule network is translation equivariant, which presents a competitive learning capacity considering the relative pose and position information of object entities in the image. The reasonable layouts of different face parts do matter during MER.
Ertugrul et al. \cite{29} manage to encode face poses and AUs at different view angles with the help of a single capsule. LaLonde et al. \cite{30} develop a deconvolutional capsule network with U-net architecture \cite{31} which achieves a good result in object segmentation. \cite{32} is the first work that manages to use capsule-based architecture for MER, whose model performance is better than the LBP-TOP baseline and several CNN models. In contrast, we propose our discriminator design which is enhanced by capsule network to implement multiple tasks: to check whether the input image is real or fake and to predict the ME labels.

\begin{figure*}
\centering
\includegraphics[width=0.85\textwidth]{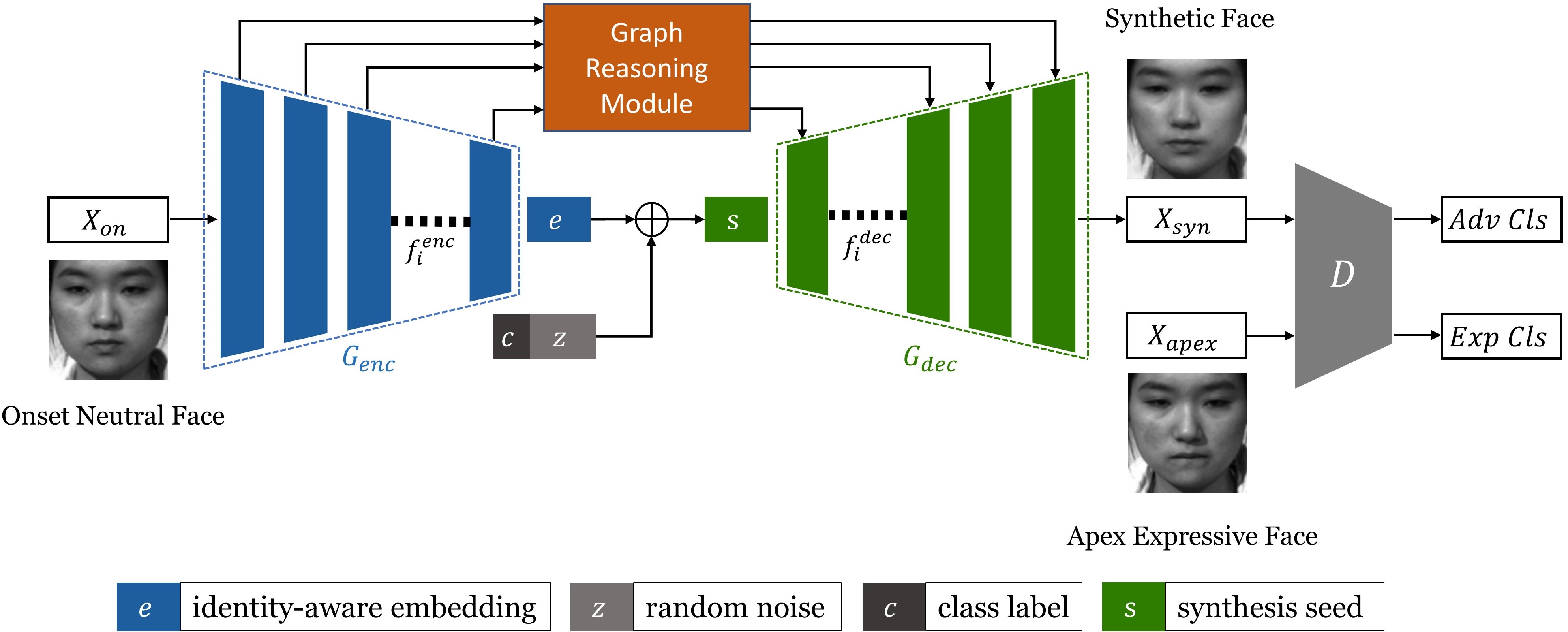}
\caption{Architecture of the proposed ICE-GAN framework for MES and MER tasks, where $G_{enc}$ and $G_{dec}$ represent the encoder and decoder part of generator, and $D$ denotes the capsule discriminator.} \label{fig:model}
\end{figure*}

\section{Methods}
The overall architecture is shown in Fig. \ref{fig:model}. The onset image $X_{on}$ is regarded as the neutral face with the lowest expressive intensity.
U-net like generator is used to modulate $X_{on}$ with side information (i.e., random noise $z$ and class label $c$) and produce the output $X_{syn}$ with a desired class, preserving the identity knowledge.
The real apex expressive faces $X_{apex}$ and $X_{syn}$ are adopted to train our capsule-enhanced discriminator, which is a multi-tasking component for authenticity checking between $X_{apex}$ and $X_{syn}$, and for expression classification.
More details are elaborated in the following sections.


\subsection{Micro-Expression Synthesis via Identity-aware Generator}
\begin{figure*}
\centering
\includegraphics[width=0.8\textwidth]{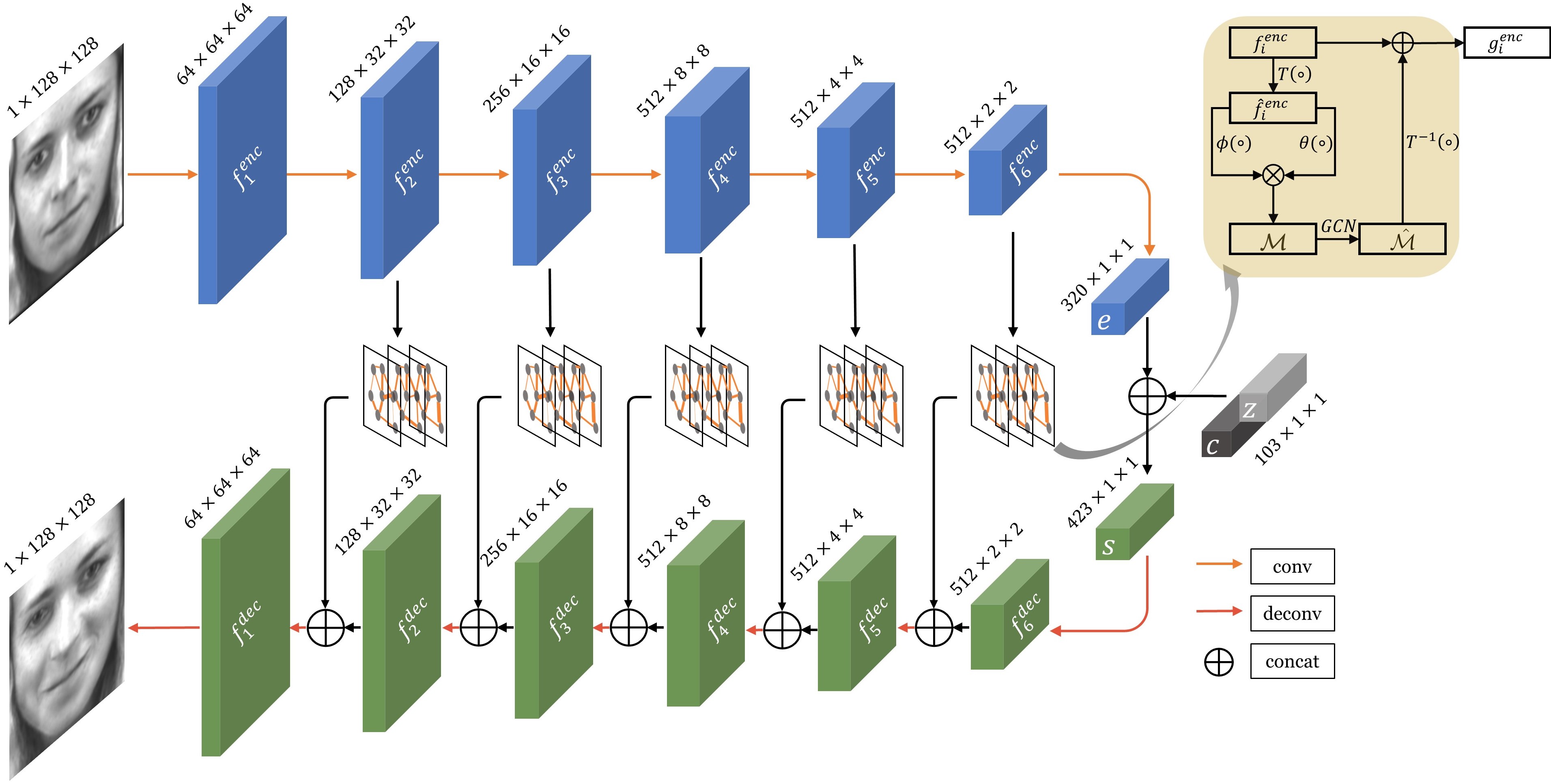}
\caption{The encoder-decoder framework of our proposed generator $G$, where the detailed implementations of the graph construction and global reasoning are illustrated within the yellow box.}\label{fig:g}
\end{figure*}

We propose an identity-aware generator $G$ based on the encoder-decoder structure shown in Fig. \ref{fig:g}, which can preserve the identity feature and produce outputs with preferable classes during MES, with the help of side information.
The encoding procedure explores facial attributes from onset neutral inputs $X_{on}$ via a series of hierarchical convolutional layers, resulting in intermediate feature maps $f_{i:1,2,...,6}^{enc} \in \mathcal{R}^{C_i \times H_i \times W_i}$ at the $i$-th convolutional layer, where $C_i$, $H_i$, and $W_i$ denote the channel, height, and width of the feature map, respectively.

Furthermore, we decide to utilize $f_{i}^{enc}$ by processing and propagating the intermediate information from the encoder $G_{enc}$ to decoder $G_{dec}$ for more realistic image generation, as these local features contain trivial knowledge such as 2D face geometries and non-trivial knowledge that is useful for the expression generation. So, we propose to use a reasoning module based on graphs, namely graph reasoning module (GRM), to capture facial part relations to reduce artifacts.

Instead of leveraging skip connections to directly transfer the multi-scale spatial information, we design to firstly flatten the spatial dimension and directly apply self-attention on feature channels to learn a channel map.
Moreover, we propose to use graph convolution to better reason the relationships between feature channels by treating the channel map as a graph, which is named as the channel graph, so that global interdependencies between intermediate facial attributes can be captured from local feature responses.

The detailed architecture is displayed on the yellow box of Fig. \ref{fig:g}.
Based on the feature map $f^{enc}_{i} \in \mathcal{R}^{C_i \times H_i \times W_i}$ after convolution, we can construct a spatial graph with the node number $N_s=H_i \times W_i$ and node feature $C_s=C_i$.
Inspired by \cite{GINet}, supernodes with richer expressiveness are formed by a transformation function $T(\cdot)$ to generate a new feature map $\hat{f}_{i}^{enc} \in \mathcal{R}^{C_i \times \hat{N_s}}$, and empirically, $T(\cdot)$ is chosen as a convolutional operation.
To learn global relations between different facial encodings, self-attention is applied to calculate a similarity mapping of features based on $\hat{f}_{i}^{enc}$, which can be formulated as:
\begin{equation}
\mathcal{M} = \phi(\hat{f}_{i}^{enc}) \theta(\hat{f}_{i}^{enc})^{T}, \label{eq:1}
\end{equation}
where both $\phi(\cdot)$ and $\theta(\cdot)$ are linear mappings.
We take the similarity mapping $\mathcal{M}\in \mathcal{R}^{C_{i} \times C_{i}}$ as a channel graph, which has a set of node features $\mathbf{m} = \{m_{j:1,2,..,C_i} \in \mathcal{R}^{1 \times C_i}\}$. In our case, we define $\mathcal{M}$ as an undirected and fully connected graph.

To reason over the whole channel graph, we utilize a GCN \cite{gcn} to learn edge features and update node features.
As the initial edge feature between two nodes is unknown and undefined, we propose to set up a self-learning adjacency matrix $A \in \mathcal{R}^{C_i \times C_i}$, which can be randomly initialized following the normal distribution and updated during back-propagation.
The whole process can be expressed as:
\begin{equation}
\hat{\mathcal{M}} = \sigma((A+I)\mathcal{M}W), \label{eq:2}
\end{equation}
where $\sigma$ is an activation function, identity matrix $I$ is added to construct self-looping, and $W\in \mathcal{R}^{C_i \times C_i}$ and $A \in \mathcal{R}^{C_i \times C_i}$ denote the weight and adjacency matrices, which are both learnable.


We then project the updated graph $\hat{\mathcal{M}}$ back to the same feature space of $f_{i}^{enc}$ through an inverse projection function $T^{-1}(\cdot)$.
Lastly, the residual connection is employed to compensate the high-level semantics relations with the low-level information.
The final output $g^{dec}_{i}$ of the graph reasoning module is represented as
$g^{dec}_{i}= f^{enc}_{i} \oplus T^{-1}(\hat{\mathcal{M}})$.

The last layer of $G_{enc}$ is the bottleneck layer that learns a compressed representation of the input data, termed as identity-aware embedding $e \in \mathcal{R}^{320 \times 1 \times 1}$.
As a controlling term for expression generation, the class label $c$ is concatenated with $e$ and random noise $z$ to form a synthesis seed $s \in \mathcal{R}^{423 \times 1 \times 1}$, which is fed into the decoding part where the spatial dimensions of feature maps are expanded and feature channels are reduced.
Thus the decoding feature maps $f^{dec}_{i:1,2,...,6}$ can be obtained with the help of deconvolution operation $h(\cdot)$ \cite{deconv}.
Moreover, $g^{dec}_{i}$ will be further concatenated with $f^{dec}_{i}$ so the learned relational information from latent feature space can be leveraged to improve synthetic image quality. The $i$-th upper layer feature map (for $i \in [2, 6]$) can be represented as:
\begin{equation}
f^{dec}_{i-1} = h(g^{dec}_{i} \oplus f^{dec}_{i}). \label{eq:4}
\end{equation}

\subsection{Micro-Expression Recognition via Capsule-Enhanced Discriminator}
Unlike CNNs operating over single scalars, capsule network attends to vectors, of which lengths are used to represent the existence probabilities of each entity in a given image.
We design a multi-tasking discriminator for sample authenticity checking and ME label classification enhanced by capsule network, which enables the learning of richer visual expressions and more sensitive to the geometric encoding of relative positions and poses of entities than conventional CNNs, dubbed a capsule-enhanced discriminator.

\begin{figure}
\centering
\includegraphics[width=0.45\textwidth]{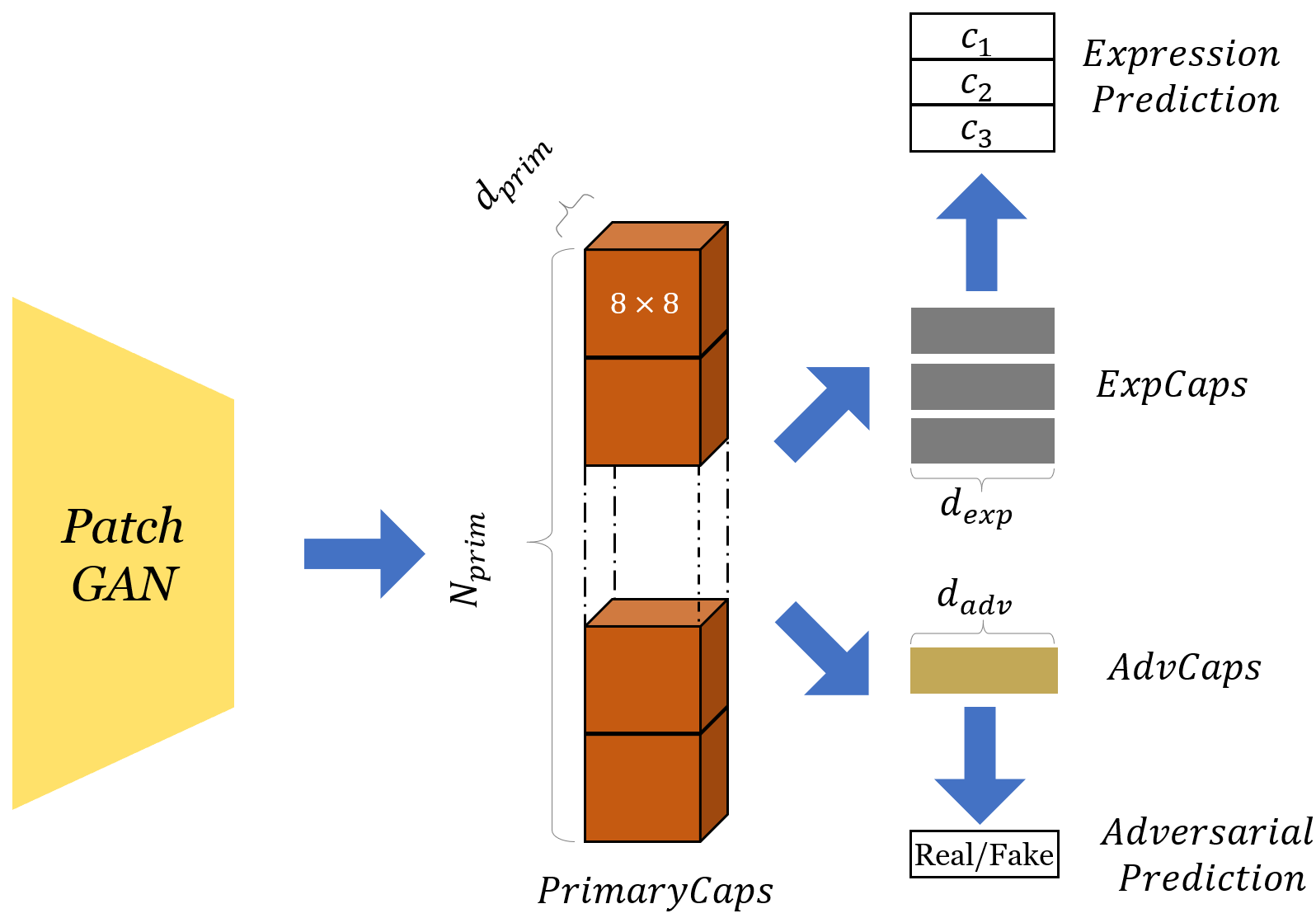}
\caption{The capsule-enhanced discriminator $D$ of our proposed method, where $N_{prim}$ denotes the number of primary capsules, $d_{prim}$, $d_{exp}$, and $d_{adv}$ represent the dimension of each \textit{PrimaryCaps}, \textit{ExpCaps}, and \textit{AdvCaps}, respectively.}
\label{fig:dis}
\end{figure}

Fig. \ref{fig:dis} presents the detailed architecture. The facial attributes are encoded via \textit{PatchGAN} for faster learning, which are then fed into \textit{PrimaryCaps} to encapsulate the information at a lower level.
Vectors generated from \textit{PrimaryCaps} are coupled and used to activate the capsules in the next layer.
Two following capsules, namely \textit{AdvCaps} and \textit{ExpCaps}, are designed for two separate sub-tasks: (1) \textit{AdvCaps} distinguishes the real expressive images $X_{apex}$ from the synthetic ones $X_{syn}$, and (2) \textit{ExpCaps} predicts the corresponding micro-expression labels of input images.
Furthermore, a reconstruction network \cite{19} is desired for a performance gain by regularizing the training of \textit{ExpCaps}.

\subsection{Model Objective}
\textbf{Generator Objective.}
A two-term identity-preserving loss $L_{ip}$ is adopted to capture the identity information embedded in $X_{on}$ (via $G_{enc}$) and thus to synthesize distinctive samples $X_{syn}$ (via $G_{dec}$). One term of $L_{ip}$ is selected as the pixel-wise reconstruction loss $L_{pixel}$ to improve the generated image quality. We empirically use L1 penalty over L2 penalty to establish direct supervision on $X_{syn}$. Meanwhile, we preserve a perceptual similarity between $X_{on}$ and $X_{syn}$ during MES as in \cite{43} by introducing a perceptual loss $L_{per}$, which is adopted as the second term in $L_{ip}$ to preserve the facial styles with regards to different subjects. With the help of a cost network which is usually implemented as a pre-trained CNN, $L_{per}$ can be easily estimated and minimized over high-level feature representations associated with $X_{on}$ and $X_{syn}$. The overall representation of $L_{ip}(G)$ can be described as:
\begin{equation}
L_{ip}(G) = L_{pixel}(G) + \alpha L_{per}(G).
\end{equation}

\textbf{Discriminator Objective.}
The training of the capsule-enhanced discriminator is optimized by the margin loss $L_{margin}$ as suggested in \cite{19}, which can enlarge the feature distance of different facial expressions. $L_{margin}$ can be obtained via:
\begin{equation}
\begin{split}
L_{margin}(D) \;=\;& T_k max(0, m^{+}-\lVert\boldsymbol{v_k}\rVert)^2 \; + \\ \;& \lambda_{k}(1-T_k)max(0,\lVert\boldsymbol{v_k}\rVert-m^{-})^2, \label{eq:6}
\end{split}
\end{equation}
where $T_k$ = 1 if expression class $k$ exists otherwise 0, while $m^{+}$ and $m^{-}$ are the upper and lower margins and $v_k$ is the vector output of capsules that being activated to class $k$.
Moreover, the mean square error is employed as the loss function $L_{rec}$ in the reconstruction network to regularize the training procedure.
Hence, the classification-related loss $L_{cls}$ for our capsule-enhanced discriminator is summed as:
\begin{equation}\label{loss_margin}
L_{cls}(D) = L_{margin}(D) + \beta L_{rec}(D).
\end{equation}

\textbf{Overall Objective.}
As we treat GAN as our baseline, the optimization process between the generator and discriminator is described as a min-max game \cite{cgan}. The learning objective $L_{gan}$ can be formulated as:
\begin{equation}
\begin{split}
L_{gan} = \mathop{min}\limits_{G}\mathop{max}\limits_{D}V(D,G) \;=\;& E_{x\sim p_{x}}[log(D(x|y))] \; + \\
\;& E_{z\sim p_{z}}[1-D(G(z|c))], \label{cgan}
\end{split}
\end{equation}
where $x$ and $y$ indicate the real images and the real labels, while $z$ and $c$ denote the random noise and fake labels.

Overall, the total loss function of our proposed ICE-GAN can be summarized as:
\begin{equation}
L_{tot} = \lambda_{adv}L_{gan}(D, G) + \lambda_{mes}L_{ip}(G) + \lambda_{mer}L_{cls}(D),
\label{eq:loss}
\end{equation}
where $\lambda_{adv}$, $\lambda_{mes}$, and $\lambda_{mer}$ are multi-tasking weight parameters of the proposed tasks including GAN optimization, identity-aware MES, and capsule-enhanced MER.

\section{Experiments}
In this section, we compared ICE-GAN with other state-of-the-art methods quantitatively for MER based on two evaluation protocols: cross-database evaluation and single-database evaluation. Additionally, qualitative analyses of synthetic images were reported on MES. Moreover, extensive experiments on our module designs were implemented for ablation studies.

\subsection{Dataset Descriptions}
Evaluations were conducted on the MEGC2019 cross-database benchmark, which consists of three publicly available ME datasets (\textit{SMIC}~\cite{smic}, \textit{CASME \RomanNumeralCaps{2}}~\cite{casme2}, and \textit{SAMM}~\cite{samm}), with a total of 442 ME video sequences (164, 145, and 133 clips collected from \textit{SMIC}, \textit{CASME \RomanNumeralCaps{2}}, and \textit{SAMM}, respectively) captured from 68 subjects (16, 24, and 28 subjects from \textit{SMIC}, \textit{CASME \RomanNumeralCaps{2}}, and \textit{SAMM}, respectively).
Frames in all 3 datasets are categorized into 3 classes of positive, negative, and surprise with the Leave-One-Subject-Out (LOSO) validation method.
LOSO is conducted for subject-independent evaluation, i.e., the evaluation is repeated each time for all subjects accordingly, until each subject is split alone as the 1-subject testing dataset and the remaining subjects as the training dataset.
In our case, we firstly conducted experiments on all 3 datasets for cross-database evaluation (CDE) based on the Unweighted F1-score (UF1) and Unweighted Average Recall (UAR), making a fair performance comparison as in \cite{see2019megc}. We then evaluated model performance only on \textit{CASME \RomanNumeralCaps{2}} and \textit{SAMM} for single-database evaluation (SDE) using F1-score following \cite{xie2020assisted}.

\subsection{Implementation Details}
In this section, we reveal detailed implementations of data preprocessing and model parameter settings for our design.

\textbf{Data Preprocessing.}
We firstly cropped out face regions by conducting facial landmark detection repeatedly to locate and refine the positions of 68 facial landmarks \cite{45}, ending up with a bounding box decided by the refined 68 landmarks with toolkit \footnote{\url{https://pypi.org/project/face-recognition/}}.
We then followed \cite{32} to find the onset and apex frames in \textit{SMIC} because annotated positions of onset and apex frames are only available in \textit{CASME \RomanNumeralCaps{2}} and \textit{SAMM}.
Moreover, we chose the neighboring four images around the apex image for each subject to augment the existing data, where we assumed that they all have the same expressive intensity.
Finally, these cropped face images were resized to 128$\times$128 in grayscale.

\begin{table*}
\caption{MER results on the MEGC2019 benchmark and separate datasets in terms of UF1 and UAR with LOSO cross-database evaluation.}\label{tab1}
\small
\centering
\begin{tabular}{|l|c|c|c|c|c|c|c|c|} 
\hline
\multirow{2}{*}{Method} & \multicolumn{2}{c|}{MEGR 2019} & \multicolumn{2}{c|}{SAMM} & \multicolumn{2}{c|}{SMIC} & \multicolumn{2}{c|}{CASME \RomanNumeralCaps{2}}  \\ 
\cline{2-9}
                        & UF1 & UAR                      & UF1 & UAR                 & UF1 & UAR                 & UF1 & UAR                    \\  \hline

LBP-TOP \cite{8}                         & 0.588                & 0.578               & 0.395                & 0.410               & 0.200                & 0.528               & 0.702                & 0.742                \\ \hline
CapsuleNet \cite{32}                         & 0.652                & 0.650               & 0.620                & 0.598               & 0.582                & 0.587               & 0.706                & 0.701                \\ \hline
Dual-Inception \cite{dualin}                        & 0.732                & 0.727               & 0.586                & 0.566               & 0.664                & 0.672               & 0.862                & 0.856                \\ \hline
STSTNet \cite{17}                           & 0.735                & 0.760               & 0.658                & 0.681               & 0.680                & 0.701               & 0.838              & 0.868                \\ \hline

ME-Recognizer \cite{18}                             & 0.788                & 0.782               & 0.775                & 0.715               & 0.746                & 0.753               & 0.829                & 0.820                \\ \hline
\textbf{ICE-GAN}                       & \textbf{0.874}                & \textbf{0.883}               & \textbf{0.879}                & \textbf{0.883}               & {\bfseries 0.782}                & \textbf{0.801}               & \textbf{0.895}                & {\bfseries 0.904}              \\ \hline
\end{tabular}
\end{table*}

\textbf{Model Parameters.}
The multi-tasking weight parameters in \eqref{eq:loss} are set as follows: $\lambda_{adv}$ = 0.1, $\lambda_{mes}$ = $\lambda_{mer}$ = 1. Two reweighting ratios $\alpha$ and $\beta$ are set to 0.1 and 5e-4.
We set the noise dimension $N_z$ to 100 and class label dimension $N_c$ to 3 as we only have 3 classes.
The discriminator includes a 70$\times$70 \textit{PatchGAN} followed by $N_{prim}$ = 8 \textit{PrimaryCaps} with dimension $d_{prim}$ = 16 and two sub-capsules for classification purposes ($d_{adv}$ = 256 and $d_{exp}$ = 32).
$m^{+}$, $m^{-}$, and $\lambda_{k}$ in \eqref{eq:6} are set to 0.9, 0.1, and 0.5.
Adam is selected as the optimizer to train our network, with momentums $b_1$ and $b_2$ set to 0.9 and 0.999, respectively. The learning rate is initialized as 1e-3 and decays using a cosine annealing schedule. The batch size is set to 16 with 100 training epochs. The end-to-end training procedure of ICE-GAN was implemented in Pytorch with one Nvidia RTX2080Ti GPU.

\subsection{Evaluations} \label{fid_section}
\textbf{Quantitative Results.}
We firstly evaluated our model performance with state-of-the-art methods from MEGC2019 benchmark \cite{see2019megc} using CDE. The results of UF1 and UAR are reported in Table \ref{tab1}. It can be observed that ICE-GAN outperforms approaches from MEGC2019 benchmark, where UF1 and UAR scores are improved by \textbf{10.9\%} and \textbf{12.9\%} compared to those of the winner \cite{18}.

The baseline (i.e., LBP-TOP) in Table \ref{tab1} adopts hand-crafted features and receives a lower result compared to deep learning methods. CapsuleNet \cite{32} utilizes capsule network to achieve an acceptable result, and the remaining works (i.e., \cite{17,dualin,18}) propose their designs based on optical flows. 
The superior performance of our design can be attributed to the expanded size of the database and the global relations of facial attributes which are learned on channel graphs.
Although ME-Recognizer \cite{18} captures spatiotemporal information from facial movements between onset and apex images, the intra-class information is ignored for each subject.

Our model also achieves the highest scores for individual dataset evaluation. For \textit{SAMM}, ICE-GAN overpasses ME-Recognizer by 13.4\% in UF1 and 23.5\% in UAR. The performance in \textit{CASME \RomanNumeralCaps{2}} has improved by 8.0\% and 10.2\% for UF1 and UAR respectively. However, the improvement on \textit{SMIC} is not as large as the previous two datasets because there is no clear annotation about onset or apex frames, which cannot give representative information about the subject.

We then compared our method with the latest state-of-the-art works following SDE.
The model performance of F1-score for 3-category classification is reported in Table \ref{tab:latest}.
Compared to MicroAttention \cite{wang2020micro}, since we capture the long-ranged interactions between different facial regions based on graph reasoning, our method achieves a higher F1-score with 8.5\% improvement on \textit{CASME \RomanNumeralCaps{2}} and 54.0\% on \textit{SAMM}.
MER-GCN \cite{MER-GCN} and AU-GACN \cite{xie2020assisted} are graph-based methods that reason over AU nodes, while our ICE-GAN achieves a better performance with F1-scores improved by 64.8\% and 43.9\% compared to AU-GACN, which indicates that our way of constructing graphs over intermediate feature channels can give more representative relational information between facial attributes than directly reason over AU graphs.

\begin{figure*}
 \centering
 \subfigure[]{
    \includegraphics[width=0.55\linewidth]{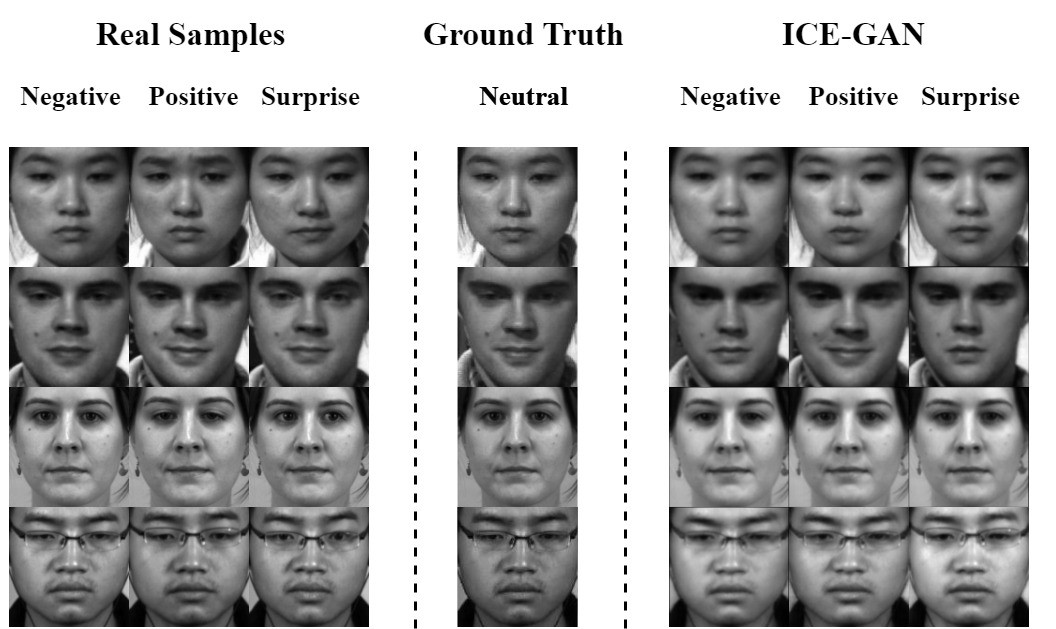}
    \label{fig:imgset_left}
 }
  \subfigure[]{
  \includegraphics[width=0.215\linewidth]{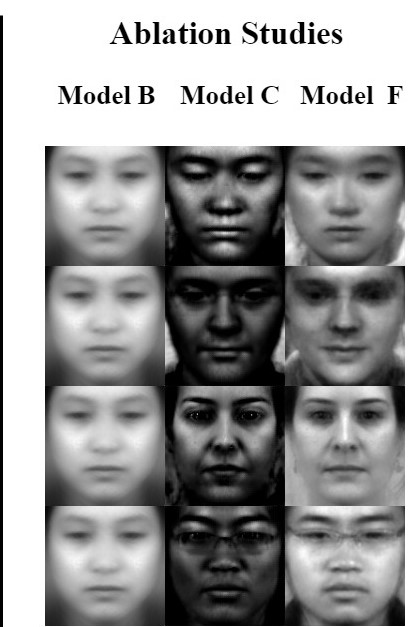}
  \label{fig:imgset_right}
 }
 \caption{(a) Comparisons between real samples and generated samples by ICE-GAN for four different subjects. Neutral images are listed for reference. (b) Synthetic images $X_{syn}$ generated by different model designs.}
 \label{fig:imgset}
\end{figure*}

\textbf{Qualitative Results.}
Furthermore, qualitative results for MES are shown in Fig. \ref{fig:imgset_left}, where comparisons between synthetic and real images of four different subjects are demonstrated. The results indicate that the generated faces achieve no significant artifacts and are basically at the same level as real samples regarding the authenticity, where light conditions are preserved in the images as well.

\begin{figure}
\centering\includegraphics[width=0.63\linewidth]{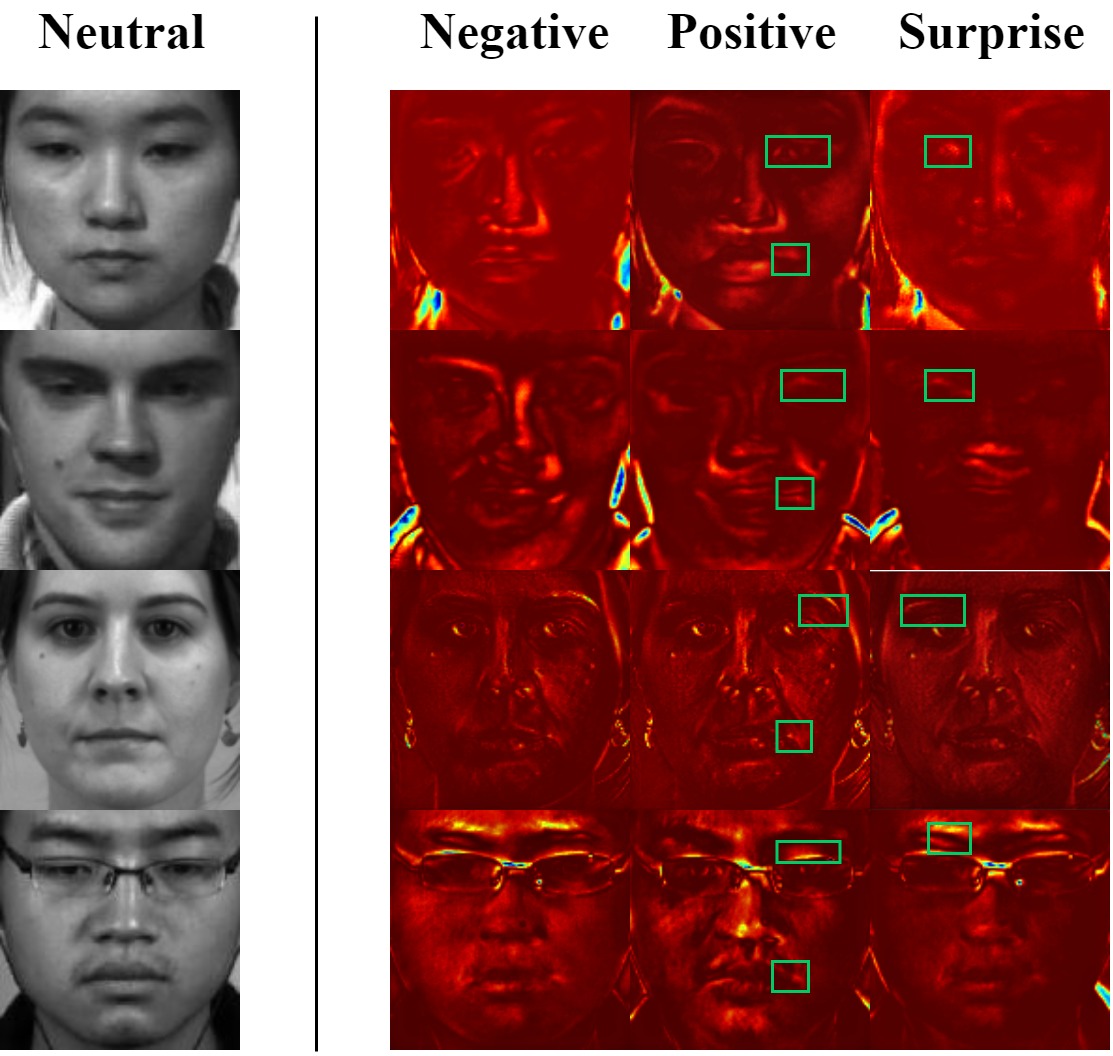}
\caption{Norm2-based difference between $X_{syn}$ and $X_{on}$. Green boxes indicate the subtle muscle movements associated with action units.} \label{fig:diff}
\end{figure}

\begin{table}
\small
\centering
\caption{Comparisons with the latest graph-based and attention-based approaches in terms of F1-score with LOSO single-database evaluation.} \label{tab:latest}
\begin{tabular}{|l|c|c|}
\hline
Method                          & CASME \RomanNumeralCaps{2}        & SAMM         \\ 
\hline
MicroAttention \cite{wang2020micro}           & 0.539                            & 0.402       \\
\hline
MER-GCN\cite{MER-GCN}           & 0.303                             & 0.283        \\
\hline
AU-GACN\cite{xie2020assisted}   & 0.355                             & 0.433        \\
\hline
\textbf{ICE-GAN}              & \textbf{0.585}                    & \textbf{0.623} \\
\hline
\end{tabular}
\vspace{-2.5mm}
\end{table}

To better examine synthetic samples, we implemented Norm2-based difference between $X_{syn}$ and $X_{on}$ and focused on AU-related face regions.
Fig. \ref{fig:diff} presents activated regions shown in red boxes for positive and surprising faces, while negative expressions have more than one kind of MEs and therefore their patterns are difficult to be visualized explicitly.
Practically, facial expressions relate to many muscle movements, so multiple AUs can be activated simultaneously.
For positive classes, green boxes drawn around the eye area indicate the activation of AU6, and the ones around the lip corner indicate AU12 and AU25.
For surprising samples, AU1, AU2, and AU5 are activated, which are all prototypical AUs that are critical for the recognition of positive (i.e., happy) and surprise micro-expressions, according to FACS \cite{6}.

\subsection{Ablation Studies}
Extensive experiments were implemented to validate the component design of ICE-GAN. The following studies were examined on the full dataset including \textit{SMIC}, \textit{CASME \RomanNumeralCaps{2}}, and \textit{SAMM}, which was further randomly split into a training set with 48 subjects and a testing set with 20 subjects.

\begin{table}
\small
\centering
\caption{Experimental analyses of $G$ and GRM designs on the full dataset split in a subject-wise manner. $SC$, $SE$, and $GR$ represent the skip connection, squeeze-and-excitation, and graph reasoning.}
\label{tab:ablation_all}
\begin{tabular}{l|p{0.3cm}|p{0.4cm}p{0.55cm}|p{0.3cm}p{0.3cm}p{0.4cm}|ll} 
\hline
Model & $D$          & $G_{dec}$    & $G_{enc}$    & $SC$         & $SE$         & $GR$          & UAR          & UF1    \\ 
\hline
A     & $\checkmark$ &              &              &              &              &               & 0.425        & 0.423  \\
B     & $\checkmark$ & $\checkmark$ &              &              &              &               & 0.651        & 0.660  \\
C     & $\checkmark$ & $\checkmark$ & $\checkmark$ &              &              &               & 0.705        & 0.707  \\
D     & $\checkmark$ & $\checkmark$ & $\checkmark$ & $\checkmark$ &              &               & 0.717        & 0.724  \\
E     & $\checkmark$ & $\checkmark$ & $\checkmark$ &              & $\checkmark$ &               & 0.719        & 0.732  \\
F     & $\checkmark$ & $\checkmark$ & $\checkmark$ &              &              & $\checkmark$ & 0.761        & 0.769  \\
\hline
\end{tabular}
\end{table}

\textbf{Analysis of Generator.}
Three models were built to verify the effectiveness of our generator in Table \ref{tab:ablation_all}.
Model \textit{A} is the baseline that just uses a discriminative model for MER.
Model \textit{B} includes a DCGAN-like generator \cite{33}, where UAR and UF1 are increased by 53.2\% and 56.0\% compared to model \textit{A}, which validates the usefulness of GAN to expand the data size.
Model \textit{C} achieves UAR of 0.705 and UF1 of 0.707 based on an encoder-decoder like structure, which proves the advantage of encoding expressive representations of the input data.

\textbf{Analysis of Graph Reasoning Module.}
To validate the excellence of the graph-based reasoning module, we compared its performance with skip connections (model \textit{D}) and squeeze-and-excitation (SE) module \cite{senet} (model \textit{E}) in Table \ref{tab:ablation_all}.
Skip connections simply propagate low-level multi-scale spatial information from encoder to decoder side, however, there is no relation learning or reasoning on latent feature space. Thus, model \textit{D} can only achieve UAR of 0.717 and UF1 of 0.724.
Results shown in model \textit{E} indicate that SE module helps the model modulate the channel interdependencies, and model capability is increased compared to model \textit{D} with only skip connections. 
Moreover, by constructing graphs based on feature channels and learning global relations on local facial features, the model performance has been further improved in the final design (model \textit{F}). Our model achieves the best performance with UAR of 0.761 and UF1 of 0.769, which indicates that relational information between different facial features reasoned from local responses contributes significantly when generating unseen faces.

\textbf{Analysis of Synthesis Quality.}
Moreover, synthetic image qualities from models \textit{B}, \textit{C}, and \textit{F} are visualized in Fig. \ref{fig:imgset_right}. The differences between models \textit{B} and \textit{C} demonstrate that finer identity-related attributes can be preserved via the encoder-decoder architecture. With the help of graph-based global reasoning, more high-frequency signals can be passed smoothly through multi-scale connections (e.g. light condition) and artifacts are largely reduced.

\textbf{Analysis of Discriminator Design.}
We then exploited the impact of the dimension $d_{exp}$ of \textit{ExpCaps} on MER, within a range of [8, 16, 32, 64, 128]. As observed in Table \ref{tab:discriminator}, we obtained the best performance when setting $d_{exp}$ to 32, with UAR of 0.761 and UF1 of 0.769.
Besides, we compared the capsule-enhanced design with its CNN-based counterparts by replacing the two-layer capsule network with a two-layer CNN, while maintaining the architecture of \textit{PatchGAN}. The performance of CNN-based discriminator is reported in the first row in Table \ref{tab:discriminator}.
To make a fair comparison of the difference in terms of model size, we increased the number of neurons in the two-layer CNN and examined the performance of the enlarged CNN-based discriminator (second row) on the MER task.
As reported in Table \ref{tab:discriminator}, our capsule-enhanced discriminator outperforms its CNN counterparts by a large margin in terms of UAR and UF1, validating that translation equivariance introduced by capsule helps MER.

\begin{table}
\small
\centering
\caption{Ablation studies on the discriminator design. $D_{CNN}$ and $D_{CE}$ denote CNN-based and capsule-enhanced discriminators.}
\label{tab:discriminator}
\begin{tabular}{l|c|c|c}
\hline
Method & UAR & UF1 & \#Params\\ \hline
$D_{CNN}$                                       & 0.356 & 0.321 & 6.7 MB\\
$D_{CNN}$ with comparable size                  & 0.432 & 0.455 & 85.6 MB\\
$D_{CE}$ with $d_{exp}=8$     & 0.730 & 0.740 & 85.7 MB\\ 
$D_{CE}$ with $d_{exp}=16$    & 0.704 & 0.723 & 85.9 MB\\ 
$D_{CE}$ with $d_{exp}=32$    & \textbf{0.761} & \textbf{0.769} & 86.4 MB\\ 
$D_{CE}$ with $d_{exp}=64$    & 0.726 & 0.732 & 87.5 MB\\ 
$D_{CE}$ with $d_{exp}=128$    & 0.713 & 0.730 & 89.6 MB\\ 
\hline
\end{tabular}
\vspace{-2.5mm}
\end{table}

\section{Conclusion}
In this work, we propose ICE-GAN which consists of an identity-aware generator with global graph reasoning on local channel graphs for micro-expression synthesis and a capsule-enhanced discriminator for micro-expression recognition.
We design a generator that encodes distinguishable facial attributes with side information to control expressions and synthesizes realistic samples by a graph reasoning module, where a channel graph is established via self-attention and global relations between long-ranged facial features are captured.
Furthermore, we present a discriminator which improves recognition ability by capturing part-based position-specific face characteristics.
Experiments on MEGC2019 datasets indicate the proposed method outperforms the winner by a significant margin, and ablation studies demonstrate our contributions to different module designs.

{\small
\bibliographystyle{ieeetr}
\bibliography{main}
}

\end{document}